\documentclass[]{ceurart}

\usepackage{listings}
\usepackage{subcaption} 
\usepackage{url} 

\begin{document}

\copyrightyear{2022}
\copyrightclause{Copyright for this paper by its authors.
  Use permitted under Creative Commons License Attribution 4.0
  International (CC BY 4.0).}
\conference{AICS-24: 32nd Irish Conference on Artificial Intelligence and Cognitive Science, December 9-10, 2024, Dublin, Ireland}

\title{Extending TWIG: Zero-Shot Predictive Hyperparameter Selection for KGEs based on Graph Structure}

\author[1]{Jeffrey Sardina}[%
orcid=0000-0003-0654-2938,
email=Jeffrey.Sardina@gmail.com,
]
\fnmark[1]
\address[1]{Trinity College Dublin, the University of Dublin, College Green
Dublin, Ireland}
\author[1]{John D. Kelleher}[%
orcid=0000-0001-6462-3248,
email=John.Kelleher@tcd.ie,
]
\author[1]{Declan O'Sullivan}[%
orcid=0000-0003-1090-3548,
email=Declan.OSullivan@tcd.ie,
]
\cortext[1]{Corresponding author.}

\begin{abstract}
Knowledge Graphs (KGs) have seen increasing use across various domains -- from biomedicine and linguistics to general knowledge modelling. In order to facilitate the analysis of knowledge graphs, Knowledge Graph Embeddings (KGEs) have been developed to automatically analyse KGs and predict new facts based on the information in a KG, a task called ``link prediction". Many existing studies have documented that the structure of a KG, KGE model components, and KGE hyperparameters can significantly change how well KGEs perform and what relationships they are able to learn. Recently, the Topologically-Weighted Intelligence Generation (TWIG) model has been proposed as a solution to modelling how each of these elements relate. In this work, we extend the previous research on TWIG and evaluate its ability to simulate the output of the KGE model ComplEx in the cross-KG setting. Our results are twofold. First, TWIG is able to summarise KGE performance on a wide range of hyperparameter settings and KGs being learned, suggesting that it represents a general knowledge of how to predict KGE performance from KG structure. Second, we show that TWIG can successfully predict hyperparameter performance on unseen KGs in the zero-shot setting. This second observation leads us to propose that, with additional research, optimal hyperparameter selection for KGE models could be determined in a pre-hoc manner using TWIG-like methods, rather than by using a full hyperparameter search.
\end{abstract}


\maketitle

\section{Introduction and Preliminaries}
Knowledge Graphs (KGs) are graph-based databases that model information as a set of nodes, which represent concepts, and edges, which represent the relationships between them \cite{kg-overview}. Knowledge Graph Embedding (KGE) models learn to predict new facts based on the information contained in a knowledge graph -- formally, this is called the link prediction task. \cite{kge-survey,rml-review,kges-for-lp-compare}. As a result of their success in link prediction, KGE models have become increasingly used in a large variety of domains -- from modelling health sciences data \cite{topological-imbalance,kges-for-cancer,kges-for-covid,kges-for-drugs,kges-for-drugs-2} to general knowledge \cite{light-into-the-dark,old-dog-new-tricks}. 

While previous studies provided detailed benchmarking of various KGE models \cite{light-into-the-dark,old-dog-new-tricks,kges-for-lp-compare,baselines-strike-back,baselines-strike-back-2}, explored the effects of specific KGE model components \cite{old-dog-new-tricks,light-into-the-dark,neg-samp-analysis,loss-func-analysis}, and explored the effects of KG structure on learning \cite{topological-imbalance,kge-poisoning,kge-poisoning-2}, no study known to the authors has attempted to create a system in which KGE models, model components, graph structure, and link prediction performance can be understood as part of a common analytic framework. In each of these areas, analysis of KGEs remain incompletely characterised in terms of the others. For example, the manner by which different KGE model components affect the learnability of various graph (sub-)structures has not been explored in detail in the literature known to the authors.

However, recent developments in the Topologically-Weighted Intelligence Generation (TWIG) approach for analysing KGE models have opened the door to characterising KGE models, KG structure, and link prediction performance in a common framework \cite{twig}. In this work, we extend the TWIG model to simulate KGEM output on multiple KGs at the same time. We provide an empirical analysis of the accuracy this new TWIG model and show that it can accurately predict the overall performance of KGE models even on previously unseen KGE hyperparameter settings. Finally, we show that it can further predict hyperparameter preference and KGE model performance on entirely unseen knowledge graphs; i.e. in the zero shot setting.

The following sections provide a background and motivation for this work. All code can be found at \url{https://anonymous.4open.science/r/TWM-4D1F/README.md}, and data files can be found at \url{https://figshare.com/s/13dc93087c97cbf7cca1}.

\subsection{Knowledge Graphs and Knowledge Graph Embeddings}
Knowledge Graphs represent data as atomic facts (also called ``triples") consisting of labelled nodes and the directed, labelled relations that occur between them. Triples in a KG are denoted as $(s,p,o)$, where $s$ represents the subject (or ``head") node, $o$ represents the object (or ``tail") node, and $p$ is the ``predicate" that describes the relationship between $s$ and $o$ \cite{kg-overview}. The intrinsically networked nature of Knowledge Graphs leads them to very naturally represent a variety of real world data, from biological pathways and biomedical data \cite{umls,primekg,bio2rdf} to linguistics \cite{fb15k237-and-wn18rr} and general knowledge \cite{yago3-10,fb15k237-and-wn18rr}. 

Knowledge Graph Embeddings are the machine learning counterpart to KGs -- they aim to automatically learn to represent all of the knowledge in a KG as latent vector embeddings of each node and edge \cite{kg-overview,kge-survey,rml-review}. These embeddings can then be used to predict new statements that should be present in a KG, allowing the inference of new knowledge from the knowledge already present in a KG -- a task called ``link prediction". The link prediction task is formally defined as answering a ``link prediction query" in the form $(s,p,?)$ or $(?,p,o)$, where $?$ represents the subject or object entity that should be predicted such that the triple would be true.

To solve this task, KGE models learn to calculate a plausibility score for all triples. This scoring function takes the form:

$$ f(e_s,e_p,e_o) \rightarrow score_{s,p,o} $$

where $e_s$ represents the embedding of the subject node, $e_p$ represents the embedding of the predicate, $e_o$ represents the embedding of the object node, and $score_{s,p,o}$ is a scalar-valued plausibility score output by the function for the given triple. 

\begin{figure}
\centering
\begin{subfigure}{0.45\textwidth}
    \includegraphics[width=\textwidth]{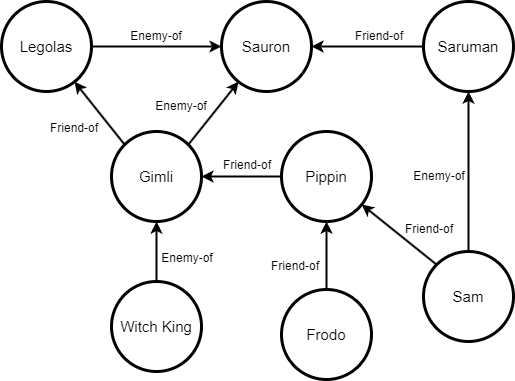}
    \caption{An example Knowledge Graph where nodes represent people and edges represent whether they are friends or enemies.}
    \label{fig:kg1}
\end{subfigure}
\hfill
\begin{subfigure}{0.45\textwidth}
    \includegraphics[width=\textwidth]{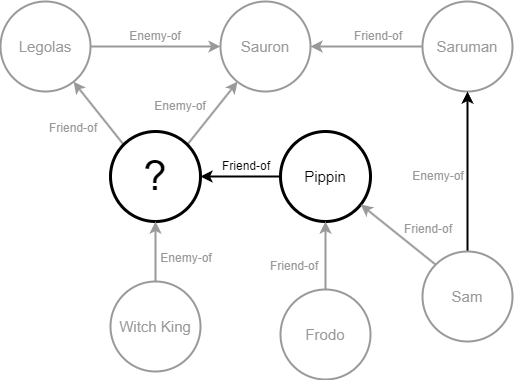}
    \caption{An example of a query for link prediction in the knowledge graph (in black) based on training data (in grey).}
    \label{fig:kg2}
\end{subfigure}
\caption{A sample Knowledge graph (left) and an example of link prediction (right).}
\label{fig:kg-examples}
\end{figure}

An example Knowledge Graph is given in Figure \ref{fig:kg1}; in this graph, all nodes represent people and all relations indicate if the people consider others to be friends or enemies. In such a graph, the link prediction task represents asking if a certain person is the friend of, or the enemy of, another. An example of link prediction in this KG is shown in Figure \ref{fig:kg2}. In this case, the query triple can be represented as $(Pippin,Friend$-$of,?)$. The remainder of the KG is not part of the query, but is used as training examples for KGE models -- in other words, it represents the background knowledge that is used by KGE models to answer the posed query.

\begin{figure}
  \centering
  \includegraphics[width=5in]{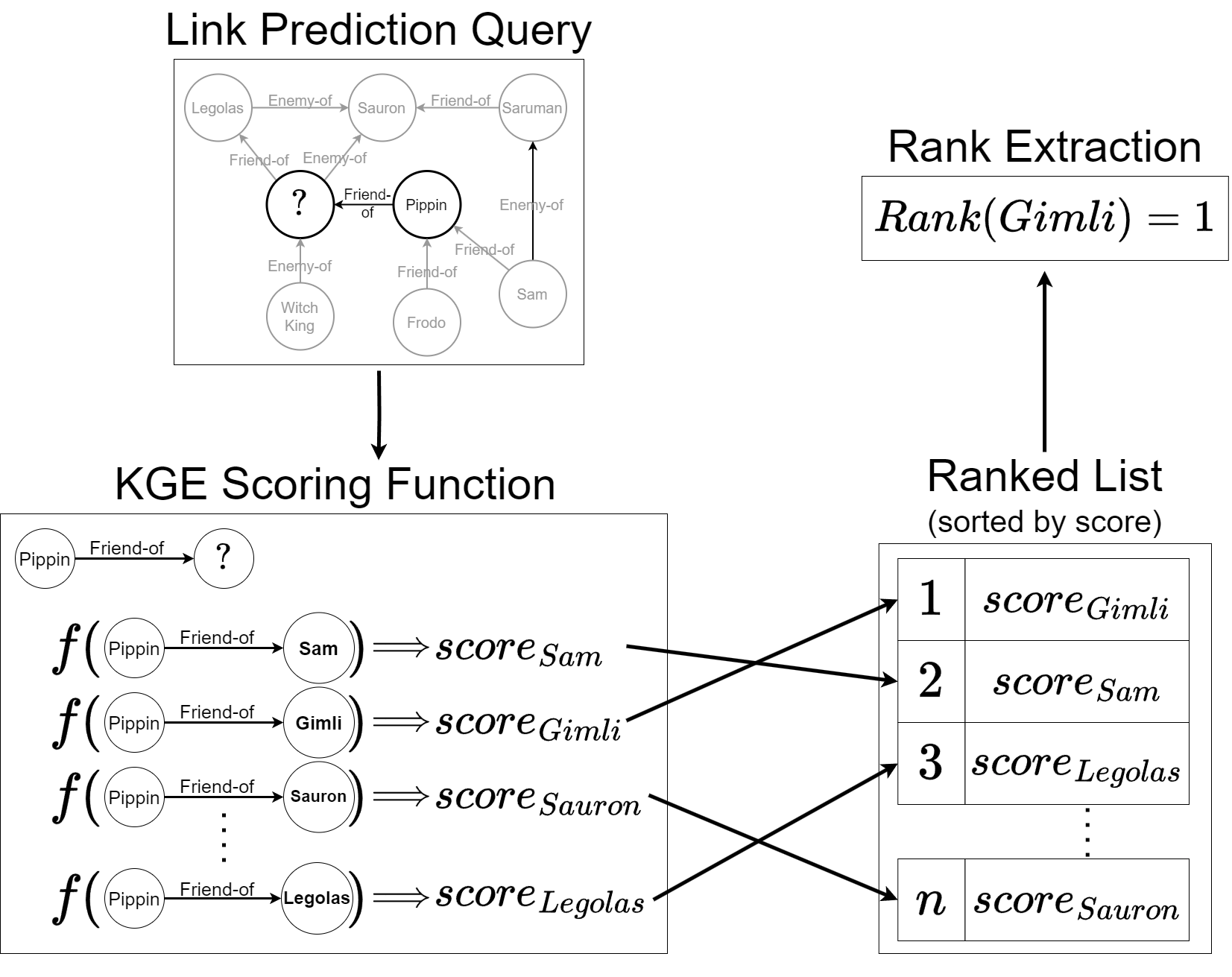}
  \caption{An schematic overview of how rank-based evaluation of KGE models is performed.}
  \label{fig:rank-evaluation}
\end{figure}

Evaluation of KGE models is based on how well the model is able to assign higher plausibility scores to known-true triples (in a KG's hold-out test set) and to assign lower plausibility scores to all other triples (not observed in the KG). A schematic overview of this procedure is shown in Figure \ref{fig:rank-evaluation}, and an algorithmic specification follows.

At an algorithmic level, in order to evaluate the performance of a KGE model on a given triple $(s,p,o)$, two ``link prediction queries" are posed based on that triple: $(s,p,?)$ and $(?,p,o)$. For each query, all possible entities in the KG are substituted for the unknown entity $?$, and all resultant triples $(s,p,\hat{o_i})$ and $(\hat{s_i},p,o)$ are scored. The scores of all link prediction queries are then sorted into two ranked lists: the list of answers to $(s,p,?)$, and the list of answers to $(?,p,o)$; both lists are sorted by score such that triples with higher plausibility scores come first in the list. The rank of the correct answers $(s^*,p,o)$ and $(s,p,o^*)$ in the sorted list is then calculated. Lower ranks (closer to 1) indicate that the correct answer to the link prediction query is predicted to be more plausible than its incorrect alternatives, whereas higher ranks (further from 1) indicate that the correct answer is predicted to be less plausible than its incorrect alternatives.

This procedure is done for all triples in the KG's hold-out test set. Once all ranks are obtained for all link prediction queries, overall performance is measured using the standard Mean Reciprocal Rank (MRR) metric \cite{light-into-the-dark,old-dog-new-tricks}. Specifically, MRR calculates the mean of the reciprocal of the ranks assigned to the correct answers of all link prediction queries. Mathematically, this is expressed as:

$$ MRR = \frac{\sum_{i=1}^{n} \frac{1}{rank_i}}{n} $$

where $n$ is the total number of link prediction queries posed and $rank_i$ is the rank of the correct answer to the $i^{th}$ link prediction query. MRR values are bounded on the interval $(0,1]$. Values closer to 1 indicate better performance, and values closer to 0 indicate worse performance.

\subsection{Properties of KGE Models}
KGE models have three main components
\begin{itemize}
  \item a \textbf{scoring function} that uses embeddings to assign a plausibility score to a triple,
  \item a \textbf{negative sampler} that produces counter-examples (in the form of fake triples) during training, and
  \item a \textbf{loss function} that forces the KGE model to assign higher scores to true triples and lower scores to ``negative" / fake triples.
\end{itemize}

On top of these, every KGE model has various hyperparameters, such as its learning rate, the size of the embeddings, its regularisation coefficients, and the number of negatives triples to generate during training. All of these hyperparameters affect how KGE learning proceeds.

Existing analysis of KGE-based link predictors has shown that the optimal choices for various KGE model components \cite{old-dog-new-tricks,light-into-the-dark,twig}, such as the hyperparameters \cite{twig,hyps-and-kg-struct}, the loss function \cite{loss-func-analysis,old-dog-new-tricks,light-into-the-dark,twig}, and the negative sampler \cite{neg-samp-analysis,old-dog-new-tricks,light-into-the-dark,twig} can be understood in terms of the KGE model in use and of the structure of the knowledge graph that is being learned.

Kotnis et al. show that the optimal negative sampler choice depends on KG structure, and suggest that this structural dependence is a significant source of hyperparameter preference in KGE models \cite{neg-samp-analysis}. However, they do note that the KGE model used also has an impact on the optimal negative sampler, citing that less expressive KGE models such as TransE cannot always benefit from more robust negative sampling protocols \cite{neg-samp-analysis}. In a similar vein, Sameh et al. show that optimal loss function choice varies both by the KGE model used and by the KG being learned \cite{loss-func-analysis}. Ruffinelli et al. and Ali et al. both conducted mass KGE benchmark studies, performing ablations of a wide variety of KGE model components on various KGs of differing structures \cite{light-into-the-dark,old-dog-new-tricks}. Overall, Ruffinelli et al. and Ali et al. both found evidence of complex systems of preference for various KGE components and KGE hyperparameters when learning different KGs, but did relatively little analysis on those systems or what gave rise to them \cite{light-into-the-dark,old-dog-new-tricks}.

Finally, a distinct body of works has explored the effectiveness of KGE models directly as a function of graph structure \cite{topological-imbalance,kges-for-lp-compare,twig,hyps-and-kg-struct}. Bonner et al. showed that so-called ``super-hub" nodes of extremely high degree could substantially impact learning, often leading to worse performance in link prediction by being over-represented in model predictions \cite{topological-imbalance}. They also highlight that nodes with very low degree (near 1, meaning that the node only connects to a select few other nodes in the graph) are much harder to learn during link prediction \cite{topological-imbalance}.

Rossi et al. showed that KGE performance can be modelled as a function of the properties of nodes and relationships -- such as the how often certain nodes and relationships co-occur in the graph \cite{kges-for-lp-compare}.

Finally, Sardina et al. showed that the rank assigned to link prediction queries during evaluation, as well as the overall performance of KGEs, can be accurately modelled using a simulation method called Topologically-Weighted Intelligence Generation (TWIG) \cite{twig}. While these results are only evaluated on a single dataset, they show a notable ability to summarise hyperparameter preference and KG learnability using a simple neural network model \cite{twig}.

Taken together, these results suggest that the performance of KGEs can be understood largely as a function of their hyperparameters / components and the structural properties of the knowledge graphs that they learn. The result of this is the theoretical possibility of predicting hyperparameter preference and KGEM performance not only in the supervised setting, but also in the few-shot and zero-shot settings -- in other words, to perform predictive hyperparameter optimisation using TWIG as a substitute to a hyperparameter search.

\section{Methodology}
The methodology of our work can be divided into two parts -- training and evaluating TWIG in a multi-KG setting, and how we define the few-shot and zero-shot evaluations for TWIG.

\subsection{The TWIG Model}
\begin{figure}
  \centering
  \includegraphics[width=5in]{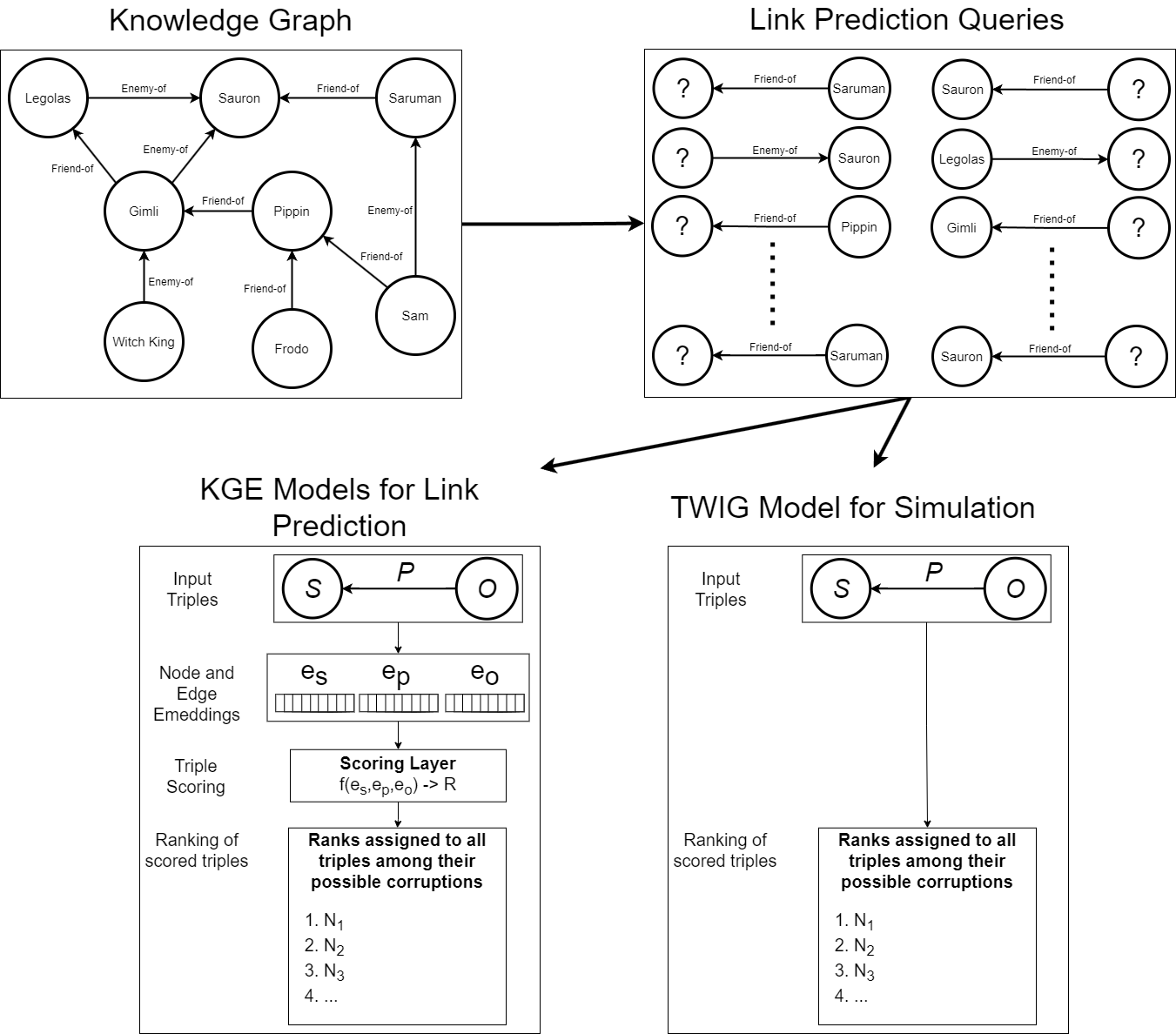}
  \caption{An overview of link prediction (using KGEs) and of KGE simulation (using TWIG).}
  \label{fig:lp-vs-simulation}
\end{figure}

\begin{figure}
  \centering
  \includegraphics[width=3.5in]{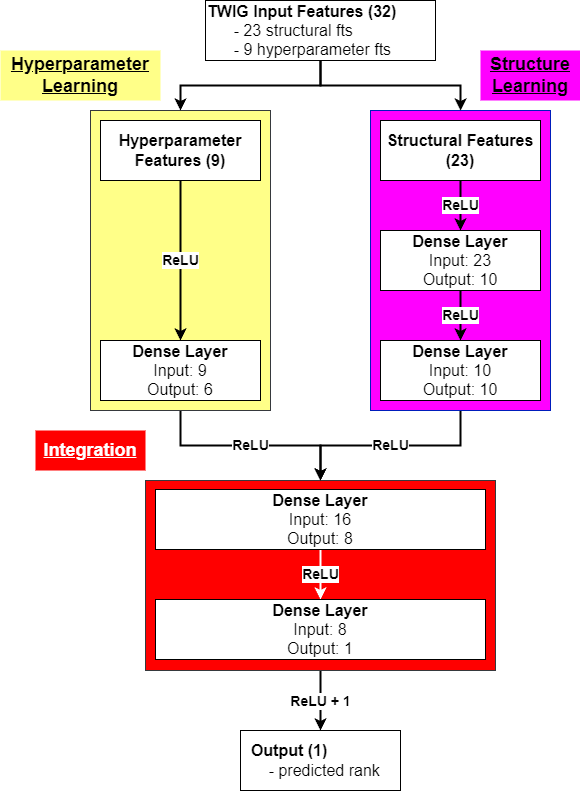}
  \caption{A visualisation of the TWIG neural network architecture. KG structure (in magenta) and hyperparameter influence (in yellow) are first learned in separate blocks, then combined in an integration component (in red) before predicted ranks are output.}
  \label{fig-twig-nn-architecture}
\end{figure}

The job of TWIG is to simulate a single knowledge graph embedding model (such as ComplEx \cite{complex-n3}) by predicting the ranks that the KGE model would assign to link prediction queries in the form $(s,p,?)$ or $(?,p,o)$. To do this, we use the TWIG neural network as published in the original TWIG paper \cite{twig}. The TWIG neural network has three major learning components -- a hyperparameter learning component, a graph structure learning component, and an integration component. The first two components learn to implicitly represent KGE hyperparameters and graph structure respectively, and the integration component combines their information to produce the final output of predicted ranks. An overview of the TWIG simulation pipeline is given in Figure \ref{fig:lp-vs-simulation}, and the architecture of the TWIG neural network is given in Figure \ref{fig-twig-nn-architecture}.

In terms of input, TWIG gathers the values of all hyperparameters, as well as the specific negative sampler and loss function being used, directly as input. It further collects fine-grained structural information on every triple in the KG, as well as aggregate statistics on each triple's local neighbourhood. Full details on all hyperparameter and structural features used by TWIG can be found in Table \ref{tab:twig-fts}; note that these are the same features used in the original TWIG paper \cite{twig}.

\begin{figure}
  \centering
  \includegraphics[width=3.5in]{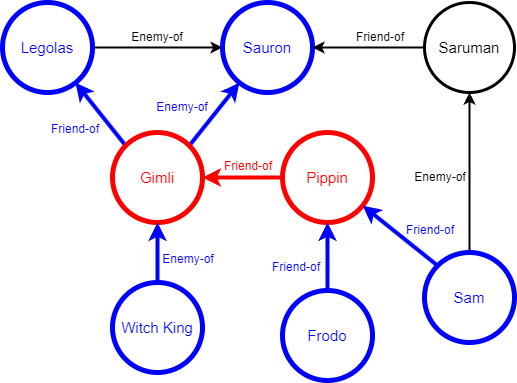}
  \caption{A visualisation of the region around a triple that TWIG uses to calculate structural features. Details statistics are gathered for the main triple (shown in red), and aggregate statistics are gathered for all triples connecting to that triple (shown in blue). Other triples (shown in black) are ignored.}
  \label{fig-fts}
\end{figure}

For simplicity, a visual depiction of the fine-grained and coarse-grained structural information that TWIG collects is shown in Figure \ref{fig-fts}. As a final note, we highlight that all structural features are calculated based only on the relations present in the KG's training set to avoid data leakage.

\begin{table}
    \centering
    \begin{tabular}{p{4.5cm}|l}
         \textbf{Feature}&\textbf{Meaning}\\ \hline
 \textbf{Hyperparameter Fts}&\\
         Negative Sampler
&The negative sampling strategy used\\
         \#Negatives per Positive
&The number of negatives sampled for each triple in training\\
         Loss Function
&The loss function used\\
         Margin (if applicable)
&The margin value used in loss calculation (if applicable)\\
         Learning Rate
&The learning rate for the Adam optimiser\\
         Embedding dimension
&The dimension of KGE embeddings\\
         Regularisation Coefficient&The coefficient of the regulariser\\  \hline
         \textbf{KG Structural Fts}&\\
 is\_head&True if the link prediction query is $(?,p,o)$; else false\\
 s\_deg&The degree of the subject node\\
 o\_deg&The degree of the object node\\
 p\_freq&
The frequency of the predicate\\
 s-p cofreq&The number of times the given subject and predicate co-occur\\
 o-p cofreq&The number of times the given object and predicate co-occur \\
 s-o cofreq&The number of times the given subject and object co-occur\\
 s min deg neighbour&The degree of the lowest-degree neighbour of the subject\\
 s max deg neighbour&The degree of the highest-degree neighbour of the subject\\
 s mean deg neighbour&The degree of the mean-degree neighbour of the subject\\
 o min deg neighbour&The degree of the lowest-degree neighbour of the object\\
 o max deg neighbour&The degree of the highest-degree neighbour of the object\\
 o mean deg neighbour&The degree of the mean-degree neighbour of the object\\
 s num neighbours&The total number of neighbours the subject has\\
 o num neighbours&The total number of neighbours the object has\\
 s min freq edge&The frequency of the least-frequent edge linked to the subject\\
 s max freq edge&The frequency of the most-frequent edge linked to the subject\\
 s mean freq edge&The mean frequency of edges linked to the subject\\
 o min freq edge&The frequency of the least-frequent edge linked to the object\\
 o max freq edge&The frequency of the most-frequent edge linked to the object\\
 o mean freq edge&The mean frequency of edges linked to the object\\
 s num edges&The total number of edges incident on the subject\\
 o num edges&The total number of edges incident on the object\\
    \end{tabular}
    \caption{A description of all features used in the TWIG model.}
    \label{tab:twig-fts}
\end{table}

\subsection{Data and Datasets}
We selected five KGs to use -- CoDExSmall, DBpedia50, Kinships, OpenEA, and UMLS \cite{codex,dbpedia50,kinships,openea,umls}, all taken from the publicly-available PyKEEN repository \cite{pykeen}. We choose these graphs for two reasons. First, they represent a diverse set of domains -- notably, biology (UMLS \cite{umls}), family trees (Kinships \cite{kinships}), and general knowledge (CoDExSmall, DBpedia50, and OpenEA \cite{dbpedia50,codex,openea}). Second, all datasets are relatively small, meaning that performing large hyperparameter benchmarking experiments to produce ground-truth data for TWIG could be done in a feasible amount of time.

In this work, we choose to simulate the state-of-the-art KGE model ComplEx, as it is among the strongest KGE models and has consistent use in many applications \cite{complex-n3,baselines-strike-back-2,light-into-the-dark,old-dog-new-tricks}. To gather information on ComplEx's performance for TWIG, we train ComplEx on a large hyperparameter grid on all five KGs, and record the ranks it assigns to all link prediction queries, as well as the overall MRR ComplEx achieves on each KG, for all hyperparameter settings. This is done a total of four times, to produce 4 replicates (differing only by random initialisations) of ranked lists to simulate for each KG and each hyperparameter combination.

We use the same hyperparameter grid as used in the original TWIG paper for training ComplEx; that grid is shown in Table \ref{tab:hp-grid}. The meaning of each hyperparameter can be found in Table\ref{tab:twig-fts}.

\begin{table}
    \centering
    \begin{tabular}{l|l}
         \textbf{Hyperparameter}& \textbf{Values Searched}\\ \hline
         Negative Sampler& Basic, Bernoulli, Pseudo-Typed\\
        \#Negatives per Positive& 5, 25, 125\\
         Loss Function& Margin Ranking, Binary Cross Entropy, Cross Entropy\\
         Margin (if applicable)& 0.5, 1, 2\\
         Learning Rate& 1e-2, 1e-4, 1e-6\\
         Embedding dimension& 50, 100, 250\\
         Regularisation Coefficient& 1e-2, 1e-4, 1e-6\\
         Epochs & 100 (constant) \\
    \end{tabular}
    \caption{The grid of hyperparameters used to train the KGE model ComplEx simulated by TWIG.}
    \label{tab:hp-grid}
\end{table}

\subsection{Evaluation of TWIG}
We evaluate TWIG in two settings: simulation of KGE performance on unseen hyperparameters, and simulation of KGE performance on unseen KGs and on unseen hyperparameters simultaneously.

\subsubsection{Evaluation of TWIG on Unseen Hyperparameters}
When evaluating TWIG on unseen hyperparameters, we define our hold-out test set as a random 10\% of all hyperparameter combinations for each KG TWIG is trained on. This is done such that the exact same hyperparameter combinations are held out on each different KG TWIG is trained on, so that when it is tested, it is tested on hyperparameter combinations it has never seen before. The remaining 90\% of hyperparameter combinations are used as the training set.

TWIG is then trained to simulate the output of all hyperparameter combinations on all five knowledge graphs for the KGE model ComplEx. This training is run exactly as in the original TWIG paper; that is, training is done in two phases, with 5 epochs for the first phase and 10 epochs for the second phase \cite{twig}. In the first phase TWIG learns using KL-divergence loss alone to teach it to match the expected distribution of ranks in its output; in the second phase TWIG learns using both KL-divergence loss and Mean Squared Error (MSE) loss to teach it to more exactly match the values of the ranks it is meant to predict while also maintaining the expected distribution of ranks \cite{twig}. Full details of this training setup, and the reasons behind 2-phase training, can be found in the TWIG paper \cite{twig}.

After training TWIG, we evaluate performance by its ability to correctly predict the Mean Reciprocal Rank (MRR) value that the KGE model ComplEx would achieve on each KG under each hyperparameter setting in the hold-out test set. To do this, we first use all of TWIG's predicted ranks for each hyperparameter combination to produce a predicted MRR score. We then calculate the $R2$ metric between ground-truth MRR values and TWIG's predicted MRR values. Note that the $R2$ metric is bounded on $(-\infty, 1]$, with higher values (closer to 1) indicating better performance and lower values indicating worse performance.

\subsubsection{Evaluation of TWIG on Unseen KGs and Unseen Hyperparameters}
When evaluating TWIG on unseen KGs, we define one of the five KGs as a hold-out set test, such that TWIG never sees the output of ComplEx on that KG during training. This is used as the first hold-out test set. For the remaining 4 KGs, we again remove a random 10\% of the hyperparameter combinations as a second hold-out test set.

We then train on the training set of the remaining 4 KGs, which consists of 90\% of the hyperparameter combinations for each KG. In order to evaluate how well TWIG can simulate the performance of ComplEx on unseen KGs, we evaluate TWIG on \textit{all} hyperparameter combinations of the hold-out KG -- which includes hyperparameter combinations it has previously seen in training, as well as hyperparameter combinations never seen during training. Once again, evaluation is done in terms of the $R2$ metric, this time between predicted MRR for each hyperparameter setting on the hold-out KG and on the true MRR value achieved by ComplEx on the same hyperparameter settings for the hold-out KG. This evaluation setting, in which an entire KG is held out, is referred to as the ``0-shot" evaluation setting.

As a generalisation of 0-shot evaluation, we also explore how TWIG performs when it is trained on a small percent of hyperparameter combinations in the unseen KG. We phrase this as a transfer learning problem, where we use the TWIG model trained on four of the five KGs as a pretrained model, and then finetune that model onto the hold-out KG. We do this in two cases -- where TWIG can see a random 5\% of the hold-out KG during finetuning, and where TWIG can see a random 25\% of the hold-out KG during finetuning. In both cases, the remaining 95\% or 75\% of the hyperparameters are reserved as a hold-out test set. We refer to these finetuning experiments as the "5\%-shot and 25\%-shot" settings respectively.

Finally, we also evaluate TWIG's ability to simulate ComplEx on each of the four seen KGs on their 25\% hold-out hyperparameter combinations, exactly as done when testing TWIG on unseen hyperparameters only. The results of all of these tests -- on seen KGs and on and unseen KGs in the  0-shot, 5\%-shot, and 25\%-shot evaluation settings, are given in the Results section.

\section{Results and Discussion}
In this section, we first demonstrate that the TWIG model we use simulates the KGE model ComplEx with high accuracy on both unseen hyperparameter combinations and on unseen KGs. We further show that TWIG can be effectively finetuned, allowing it to be re-purposed to new KGs with minimal effort and very little data for finetuning.

\subsection{Evaluation on Unseen Hyperparameters}
Table \ref{tab:twig-res} gives the results of TWIG when trained on 90\% of all hyperparameter combinations of all 5 KGs, and tested on the remaining 10\% of hyperparameter combinations on the same KGs.

\begin{table}[!ht]
    \centering
    \begin{tabular}{c|c|c|c|c|c}
        & \textbf{CoDExSmall} & \textbf{DBpedia50} & \textbf{Kinships} & \textbf{OpenEA} & \textbf{UMLS} \\ \hline
        TWIG $R^2$ & 0.95 & 0.72 & 0.98 & 0.8 & 0.96 \\
    \end{tabular}
    \caption{$R2$ values achieved by TWIG for each KG when trained jointly on the training sets of all KGs. $R2$ is calculated between observed and predicted MRR values for each hyperparameter combination on each KG.}
    \label{tab:twig-res}
\end{table}

These results indicate that TWIG can generalise across various KGs and effectively simulate the outputs of KGEs on many different hyperparameter settings and many different KGs. The $R2$ values of TWIG on each dataset all lie between 0.72 at the lowest (on DBpedia50) and 0.98 at the highest (on Kinships), indicating that it can accurately predict the performance of ComplEx on all KGs under various hyperparameter settings. We further highlight that as this evaluation is done on hold-out hyperparameters, that it indicates that TWIG is able to predict the efficacy of hyperparameter combinations it has never seen.

\subsection{Evaluation on Unseen KGs}
Table \ref{tab:0-shot-res} gives all results of TWIG when trained on 90\% of all hyperparameter combinations of 4 KGs, and tested:
\begin{itemize}
    \item on the remaining 10\% of hyperparameter combinations on the 4 KGs seen during training (shown on the left)
    \item on all hyperparameter combinations on a hold-out KG never seen during training (shown on the right)
\end{itemize}

Table \ref{tab:0-shot-res} further shows the results of finetuning TWIG with either a random 5\% (for the 5\%-shot setting) or a random 25\% (for the 25\%-shot setting) of all hyperparameter combinations on the hold-out KG before evaluating on that KG. All combinations of 4 training KGs and one hold-out KG are shown.

\begin{table*}
  \centering
    \small
  \begin{tabular}{cccc|ccc}
      \multicolumn{4}{c}{\textbf{Training KGs}} & \multicolumn{3}{c}{\textbf{Testing KG}}  \\
      \multicolumn{4}{c}{\textbf{}} &
      \multicolumn{1}{c}{\textbf{0-shot}} & \multicolumn{1}{c}{\textbf{5\%-shot}}  & \multicolumn{1}{c}{\textbf{25\%-shot}}      \\ \hline

        CodExSmall & DBpedia50  & Kinships & OpenEA &  & UMLS &  \\
        0.95      & 0.81        & 0.98       & 0.83   & 0.64 & 0.91 & 0.97 \\ \hline

        CodExSmall & DBpedia50  & Kinships & UMLS &  & OpenEA &  \\
        0.95      & 0.85        & 0.94       & 0.96   & 0.54 & 0.77 & 0.97 \\ \hline

        CodExSmall & DBpedia50  & OpenEA & UMLS &  & Kinships &  \\
        0.97      & 0.83        & 0.83       & 0.86   & 0.65 & 0.90 & 0.99 \\ \hline

        CodExSmall & UMLS  & Kinships & OpenEA &  & DBpedia50 &  \\
        0.95      & 0.93        & 0.95       & 0.88   & 0.73 & 0.81 & 0.86 \\ \hline

        UMLS & DBpedia50  & Kinships & OpenEA &  & CoDExSmall &  \\
        0.98      & 0.88        & 0.98       & 0.89   & 0.73 & 0.96 & 0.98 \\
  \end{tabular}
\caption{$R^2$ values achieved by TWIG in the on unseen hyperparameters, as well as in the 0-shot, 5\%-shot, and 25\%-shot settings on all combinations of 4 training KGs and 1 unseen KG. $R2$ is calculated between ground truth and predicted MRR values for each hyperparameter combination on each KG.}
\label{tab:0-shot-res}
\end{table*}

The results indicate that TWIG is able to simulate ComplEx not only on unseen hyperparameters, but also on entirely unseen KGs. Particularly notably, TWIG achieves an $R2$ of between 0.54 (at the lowest) and 0.73 (at the best) in the 0-shot setting where it is not trained on any information about the hold-out KG. This improves to 0.77 (at the worst) to 0.96 (at the best) in the 5\%-shot setting and 0.86 to 0.99 in the 25\%-shot setting, where 5\% or 25\% of the hold-out KG is used to finetune TWIG before evaluation.

We highlight that these results have two immediate, major implications:
\begin{itemize}
    \item that TWIG is able to use a general knowledge of KG structure and hyperparameter effects to simulate KGEs on datasets and hyperparameter combinations it has never seen in training, and
    \item that TWIG is highly receptive to finetuning, and that finetuning even on small amounts of data can lead to substantial increases in its ability to simulate KGEs.
\end{itemize}

We further highlight that the high performance of the 5\%-shot and 25\%-shot evaluation setting is not unexpected -- since TWIG is finetuned on only one dataset, rather than having to simulate KGEs on all datasets, it has a much higher ability to learn the details of that KG in particular. When trained on multiple KGs, TWIG's performance drops somewhat due to the need to simulate KGEs in so much more diverse of an environment. In fact, a weaker version of this effect is seen in the slight per-KG increase in performance on unseen hyperparameters observed when TWIG is trained 4 KGs only (as in Table \ref{tab:0-shot-res}), versus when it is trained on all 5 (as in Table \ref{tab:twig-res}).

Finally, it is important to highlight that zero-shot and few-show prediction both work \textit{regardless of the domain of the unseen KG}. For example, when predicting hyperparameter preference for Kinships (the only KG containing family-tree data) and UMLS (the only KG containing biological data), R2 values of 0.65 and 0.64 are obtained respectively. In both cases, no family tree data, nor biological data, was present in the KGs TWIG was trained on. This suggests that structural impact on hyperparameters may be domain-agnostic, allowing it to generalise across KGs even from different knowledge domains and with distinct information content.

\section{Conclusion}
In this paper, we extend TWIG, a novel system that can predict KGE model performance and hyperparameter preference for unseen KGs based on graph structure. Our results indicate that TWIG can accurately predict and represent the overall performance of the KGE model ComplEx on both unseen hyperparameters and on unseen KGs. This last ability, its ability to predict overall KGE performance for various hyperparameter sets on unseen KGs, is very significant for several reasons.

First, it suggests that hyperparameter preference is a function of KG structure. Since TWIG is able to make this prediction across KGs from different domains, it further highlights that structural are fundamental to KGE-based KG learning. It is therefore possible that the information content of the KG, paradoxically, may be less important than KG structure for determining how well the KG can be learned by KGE models.

Second, the substantively increased performance on formerly unseen KGs in the few-shot setting suggests that TWIG could possibly serve as a replacement for large-scale hyperparameter searches. Since TWIG can predict hyperparameter preference with high accuracy when trained on only 5\% or 25\% of a hyperparameter grid, and since this effect persists across various KGs from different domains, there is strong initial evidence that it could be used in lieu of a traditional hyperparameter search.

Finally, the ability of TWIG to model hyperparameter preference and KGE model performance across multiple KGs (seen and unseen) using their structural characteristics suggests that some elements of KG structure are common among different KGs. For example, Bonner et al. \cite{topological-imbalance} have shown that, in the presence of strongly skewed distributions of node degrees, KGE learning can be heavily biased. It is very possible that a variety of effect such as this underpin TWIG's ability to predict KGE performance. Determining the exact nature of such structural relations, and to what extent TWIG uses them in its predictions, is left for future work.

We must finally acknowledge some limitations of this study. Most notably, only one KGE model (ComplEx) has been tested here. While our results clearly show that TWIG can generalise across KGs, it remains unknown if it can generalise across different KGE models. We leave testing other literature-standard models, such as DistMult and TransE, to future directions.

Further, all KGs examined in this work are relatively small, especially compared to the standard benchmark KGs FB15k-237 and WN18RR. It is, as such, unclear how TWIG reacts to KGs of different size but otherwise similar structure, or if TWIG can effectively generalise across KGs of much more variable size. Evaluating TWIG in these settings is also left as a future direction.

\section*{Acknowledgements}
This research was conducted with the financial support of Science Foundation Ireland D-REAL CRT under Grant Agreement No. 18/CRT6225 at the ADAPT SFI Research Centre at Trinity College Dublin, together with sponsorship of Sonas Innovation Ireland.  The ADAPT SFI Centre for Digital Content Technology is funded by Science Foundation Ireland through the SFI Research Centres Programme and is co-funded under the European Regional Development Fund (ERDF) through Grant \# 13/RC/2106\_P2.

\bibliography{article-main}

\end{document}